\documentclass{article}
\usepackage[utf8]{inputenc}
\usepackage{times}
\usepackage{epsfig}
\usepackage{graphicx}
\usepackage[caption=false,font=footnotesize]{subfig}
\usepackage{amsmath}
\usepackage{amssymb}
\usepackage{commath}

\usepackage{multirow}
\usepackage{bm}
\usepackage{comment}
\usepackage{booktabs}
\usepackage{algorithm,algpseudocode}
\usepackage{soul}
\usepackage{multirow}

\usepackage{amsmath,amsfonts,bm}

\def\eqref#1{equation~\ref{#1}}

\def\1{\bm{1}}

\def\vb{{\bm{b}}}

\def\vm{{\bm{m}}}

\def\vt{{\bm{t}}}

\def\vx{{\bm{x}}}

\DeclareMathAlphabet{\mathsfit}{\encodingdefault}{\sfdefault}{m}{sl}
\SetMathAlphabet{\mathsfit}{bold}{\encodingdefault}{\sfdefault}{bx}{n}

\newcommand{\R}{\mathbb{R}}

\DeclareMathOperator*{\argmax}{arg\,max}

\usepackage{authblk}
\usepackage{xcolor,colortbl}

\usepackage[pagebackref=true,breaklinks=true,colorlinks,bookmarks=false]{hyperref}

\DeclareMathOperator*{\Top}{\textup{Top}}

\usepackage[capitalise]{cleveref}

\title{Bit-Flip Attack: Crushing Neural Network with Progressive Bit Search}

\author[1]{Adnan Siraj Rakin and Zhezhi He \thanks{Equally contributed and Co-First Author}}
\author[1]{Deliang Fan\thanks{Corresponding Author: dfan@ucf.edu}}
\affil[1]{Department of Computer Engineering, University of Central Florida}

\date{}

\begin{document}

\maketitle

\begin{abstract}
 Several important security issues of Deep Neural Network (DNN) have been raised recently associated with different applications and components. The most widely investigated security concern of DNN is from its malicious input, a.k.a  adversarial example. Nevertheless, the security challenge of DNN's parameters is not well explored yet. In this work, we are the first to propose a novel DNN weight attack methodology called Bit-Flip Attack (BFA) which can crush a neural network through maliciously flipping extremely small amount of bits within its weight storage memory system (i.e., DRAM). The bit-flip operations could be conducted through well-known Row-Hammer attack, while our main contribution is to develop an algorithm to identify the most vulnerable bits of DNN weight parameters (stored in memory as binary bits), that could maximize the accuracy degradation with a minimum number of bit-flips. Our proposed BFA utilizes a Progressive Bit Search (PBS) method which combines gradient ranking and progressive search to identify the most vulnerable bit to be flipped. With the aid of PBS, we can successfully attack a ResNet-18 fully malfunction (i.e., top-1 accuracy degrade from 69.8\% to 0.1\%) \textbf{only through 13 bit-flips out of 93 million bits}, while randomly flipping 100 bits merely degrades the accuracy by less than 1\%.

\end{abstract}

\section{Introduction}
Recently, deep neural networks (DNNs) have demonstrated its great potential of surpassing or close to human-level performance in multiple domains, such as object recognition \cite{he2015delving}, Game AI \cite{silver2017mastering}, synthetic voice \cite{oord2016wavenet}, neighborhood voting prediction \cite{gebru2017using} and etc \cite{gatys2015neural}. It stimulates the demand for deploying the state-of-the-art deep learning algorithms in real-world applications to release labors from repetitive work. Under such circumstance, the security and robustness of deep neural network is an essential concern which cannot be circumvented.

Adversarial example \cite{goodfellow2014explaining} (aka., adversarial attack) is a well-known security issue of DNN, which can cause the system malfunction with the magnitude-constrained input noise that mankind cannot discern. Both attack and defense of adversarial example on the input end of DNN has been heavily investigated in the past couple of years \cite{madry2018towards,goodfellow2014explaining,sun2018feature} and still be in progress \cite{rakin2018parametric,prakash2018deflecting,liao2018defense}. Nevertheless, the security issue of network parameters themselves is not yet well explored. Recently, the development of fault injection attack \cite{liu2017fault} has raised further security concerns on the storage of DNN parameters. 



The possible reasons that there was a lack of concerns on the security of network parameters may come in twofold: 1) The neural network is widely recognized as a robust system against parameter variations. 2) The DNNs are used to be only deployed on the high-performance computing system (e.g., CPUs, GPUs, and other accelerators \cite{seshadri2017ambit,angizi2018cmp}), which normally contains a variety of methods ensuring data integrity. Thus, attacking the parameters is more related to a system cyber-security topic. However, the game has been totally changed during the past few years. First, the robustness of neural network to small perturbation has been put into the spotlight by adversarial examples on DNN input \cite{goodfellow2014explaining,madry2018towards}. Second, with the aid of DNN compression techniques (e.g., pruning\cite{han2015deep} and quantization \cite{zhou2016dorefa}) and outstanding compact neural network architectures \cite{iandola2016squeezenet,sandler2018mobilenetv2}, deep neural networks now are friendly to the resource-limited mobile device as well. Such resource-limited platforms normally lack effective data integrity check mechanism, which makes the deployed DNN vulnerable to popular fault injection techniques, such as row hammer and laser beam \cite{barenghi2012fault}. 

Recently, there exist a cohort of works \cite{liu2017fault,breier2018deeplaser} in an attempt to attack DNN network parameters stored in DRAM using Row Hammer Attack (RHA). However, the key limitation to these previous attack methods is that they primarily focused on extremely vulnerable full-precision DNN model (i.e., parameters in floating-point format). Our conducted simulation shows that randomly flipping the exponent part of floating-point weight could easily overwhelm the functionality of DNN. The explanation behind that is flipping the bits in exponent part of floating-point value can increase the weight to extremely large value, thus leading to the exploded output. As a result, attacking the weight constrained DNN (i.e., weights quantized into fixed-point values) is the primary focus in this work, where the range of weight magnitude relies on the bit-width of weights.


\paragraph{Overview of Bit-Flip attack:} In this work, we attempt to perform parameter attack on the weights of quantized DNN, whose weight magnitude is intrinsically constrained owing to the fixed-point representation. 
In order to conduct an efficient bit-flip attack on weights, for the first time, we propose a Bit-Flip Attack (BFA) together with Progressive Bit Search (PBS) technique, that can totally crush a fully functional quantized DNN and convert it to a random output generator with several bit-flips. Our proposed PBS combines gradient ranking and progressive search to locate the most vulnerable bits, while BFA performs the bit-flip operations on the located bits along their gradient ascending directions. In order to identify the vulnerable bits to be flipped within the identical layer and across different layers, we perform the in-layer search and cross-layer search in an iterative way. Thus, for each BFA iteration, only the most vulnerable bit elected by the PBS technique will be flip to its opposite binary value. The extensive experiments are conducted regarding various network structure, different datasets and quantization bit-width, and etc. It is shocking to notice that ResNet-18 will become a random output generator (i.e., 0.1\% top-1 accuracy) with only 13 bit-flips out of 93 million bits by our proposed attacking method, on ImageNet dataset.

\section{Related Work}

\paragraph{Memory Bit-Flip in Real-World:}

Flipping a memory cell bit within memory system is a realistic and demonstrated threat model in existing computer systems. Recently, Kim \textit{et al.}, \cite{kim2014flipping} have demonstrated a method to cause memory bit-flip in DRAM merely through the frequent data accessing, which is now popularly known as Row-Hammer Attack (RHA). A malicious user can use RHA to modify the data stored in DRAM memory cell by just flipping one bit at a time. \cite{razavi2016flip} showed that by creating a profile for the bit flips in a DRAM, row hammer attack can effectively flip a single bit at any address in the software stack. According to the state-of-the-art investigations, common error detection and correction techniques, such as Error-Correcting Code (ECC) \cite{cojocar19exploiting} and Intel SGX \cite{gruss2018another}, are broken defense mechanism to RHA. Such existing memory bit-flip attack (i.e. row-hammer attack) model brings a huge challenge to the security of DNN powered computing system since its parameters are normally stored in the main memory, i.e. DRAM, for maximizing the computation throughput, which is directly exposed to the adversarial attacker. Moreover, such challenge becomes more severe considering the fact that DNN powered applications are widely deployed in many resource-limited (e.g. smart IoT devices, mobile system, edge devices, etc.) system that lacks necessary data integrity check mechanism.

\paragraph{Previous Neural Network Parameter Attack.}
Adversarial example attack has been widely explored \cite{yuan2019adversarial} to evaluate the robustness of DNN. However, we are still at the rudimentary stage towards investigating the effect of network parameter attack on neural network accuracy. Neural network parameters have been attacked using different levels of hardware trojans, which require a specific pattern of input to trigger the trojan inside the network \cite{clements2018hardware}. Moreover, such trojan attack requires hardware level modifications, which may not be feasible in many practical applications. As a result, fault injection attacks could become a suitable alternative to attack DNN parameters \cite{liu2017fault}. For example, single Bias attack (SBA) attacks a certain bias term of a neuron to change the classification of DNN to a different class \cite{liu2017fault}. Other works have injected faults into the activation function of the neural network to miss classify a target input \cite{breier2018deeplaser}. 

\paragraph{Limitations of previous works.}

However, these previous attack algorithms are developed based on a full-precision model (i.e. network parameters are floating-point numbers stored in memory in the format of IEEE standard for floating-point arithmetic \cite{hollasch2005ieee}), where we believe such attack algorithms may not be efficient. Since it is extremely easy to cause DNN malfunction by just flipping the most significant exponent bits of any random floating-point weight parameters. Through this simple method, it mainly causes DNN malfunction by exponentially increasing the magnitude of particular weight parameters by just several bit-flips. We conducted such experiment to prove its efficiency in section \ref{randomflip}. Based on our simulation results, it shows just 1 bit-flip of the most significant exponent bit of a random floating-point number weight could cause ResNet-18 network totally malfunction on ImageNet dataset.

\paragraph{Why we need a bit search algorithm.}
On the other side, most of recent deep neural network applications are performed in quantized platform such as google's Tensor Processing Unit (TPU) \cite{wu2016google}, that uses 8-bit operations for quantized network. Such fixed precision models are more robust to network parameter perturbation. Similarly, we conducted another experiment to randomly choose quantized weight for bit-flip attack using RHA. The simulation results in figure \ref{fig:random_attack} show that 100 bit-flip in a quantized ResNet-18 could only cause 0.6\% accuracy degradation in ImageNet, which clearly indicates that random selection of quantized weight parameters to be attacked is not efficient and feasible. 
Thus, an efficient algorithm is required to search for the most vulnerable weights/bits in a quantized DNN.

\section{Approach}
In this section, we present a novel Bit-Flip Attack (BFA) method to maliciously cause a DNN system malfunction through flipping extremely small amount of vulnerable bits of weights. Our proposed algorithm, called Progressive Bit Search (PBS), is to identify those vulnerable DNN weight parameters (stored in terms of memory bits in DRAM) that could maximize the accuracy degradation with minimum number of bit-flips. It is worth to note that this work focuses on BFA on a more robust DNN with quantized weight parameters instead of floating-point number weights as discussed earlier.


\subsection{Problem Definition}
\label{prob_def}

Given a quantized DNN contains $L$ convolutional/fully-connected layers, the original weights in floating-point are symmetrically quantized into $2^{N_q}-1$ levels with $N_q$-bits uniform quantizer. The quantized weights $\textbf{W}$ are arithmetically represented in $N_q$-bits signed integer. In the computing memory system, $\textbf{W}$ is stored in the format of twos complement\footnote{All the binary weight mentioned hereinafter referred to as the weights in twos complement.}, which is denoted as $\textbf{B}$ in this work. More details of weights quantization are described in \cref{sec:quan&encode}. The goal of this work is to find the optimal combination of vulnerable weight bits to perform BFA, thus maximizing the inference loss of DNN parameterized by the perturbed weights whose twos complement representation is $\hat{\textbf{B}}$. Such vulnerable bit searching problem can be formulated as an optimization problem as:
\begin{equation}
\begin{gathered}
\max_{\{\hat{\textbf{B}}_l\}}  ~\mathcal{L}\Big (f \big( \vx ; \{\hat{\textbf{B}}_l\}_{l=1}^{L} \big), \vt \Big) - \mathcal{L}\Big (f \big( \vx ; \{\textbf{B}_l\}_{l=1}^{L} \big), \vt \Big) \\
\centering \textup{s.t.}  ~\small{\sum_{l=1}^{L}} \mathcal{D}(\hat{\textbf{B}}_l, \textbf{B}_l) \in \{0,1,...,N_b\}
\end{gathered}
\end{equation}
where $\vx$ and $\vt$ are the vectorized input and target output\footnote{Note that, all the targets $\vt$ in this work are not the ground-truth labels, but the outputs of the clean DNN w.r.t the input data.}. Taken $\vx$ as the input, the inference computation of network parameterized by $\{\hat{\textbf{B}}_l\}_{l=1}^{L}$ is expressed as $f(\vx;\{\hat{\textbf{B}}_l\}_{l=1}^{L})$. Note that  
$\mathcal{L}(\cdot,\cdot)$ calculates the loss between DNN output and target. $\mathcal{D}(\hat{\textbf{B}}_l, \textbf{B}_l)$ computes the Hamming distance between clean- and perturbed-binary weight tensor, and $N_b$ is maximum Hamming distance allowed through the entire DNN. 

\subsection{Quantization and Encoding}
\label{sec:quan&encode}


\paragraph{Weight quantization.}
In this work, we adopt a layer-wise $N_q$-bits uniform quantizer for weight quantization. For $l$-th layer, the quantization process from the floating-point base $\textbf{W}_l^{\textup{fp}}$ to its fixed-point (signed integer) counterpart $\textbf{W}_l$ can be described as:
\begin{equation}
\label{eqt:quan_stepsize}
    \Delta w_l = \textup{max}(\textbf{W}_l^{\textup{fp}})/(2^{N_q-1}-1); \quad \textbf{W}_l^{\textup{fp}} \in \R^d
\end{equation}
\begin{equation}
\label{eqt:quan_func}
    \textbf{W}_l = \textup{round}(\textbf{W}_l^{\textup{fp}}/\Delta w_l) \cdot \Delta w_l
\end{equation}
where $d$ is the dimension of weight tensor, $\Delta w_l$ is the step size of weight quantizer. For training the quantized DNN with non-differential stair-case function (in \cref{eqt:quan_func}), we use the straight-through estimator \cite{bengio2013estimating} as other works \cite{zhou2016dorefa}. Note that, since $\Delta w_l \in \R$ is the coefficient shared by all the weights in $l$-th layer, we only store its fixed-point part $(\textbf{W}_l/\Delta w_l) \in \{-2^{N_q-1},...,2^{N_q-1}\}^d$, rather than $\textbf{W}_l$.


\paragraph{Weight Encoding.}
The computing system normally stores the signed integer in two's complement representation, owing to its efficiency in arithmetic operations (e.g., {\tt mul}). Given one weight element $w \in \textbf{W}_l$, the conversion from its binary representation ($\vb=[b_{N_q-1},...,b_{0}
]\in \{0, 1\}^{N_q}$) in two's complement can be expressed as:
\begin{equation}
\label{eqt:twoscomplement}
w/\Delta w = g(\vb) = -2^{N_q - 1}\cdot b_{N_q-1} + \sum_{i=0}^{N_q-2} 2^{i}\cdot b_{i}
\end{equation}
With the conversion relation described by $g(\cdot)$ in \cref{eqt:twoscomplement}, we can inversely obtain the binary representation of weights $\textbf{B}$ from its fixed-point counterpart as well. 

\subsection{Bit-Flip Attack}
\label{sec:BFA}
In this work, we perform the BFA utilizing the similar mechanism as FGSM \cite{goodfellow2014explaining}, which was used to generate adversarial example. The key idea of BFA is to flip the bits along its gradient ascending direction w.r.t the loss of DNN. We take the binary vector $\vb$ in \cref{eqt:twoscomplement} as an example and attempt to perform BFA upon $\vb$. We first calculates the gradients of $\vb$ w.r.t loss as:
\begin{equation}
\label{eqt:bit_grad}
    \nabla_{\vb} \mathcal{L} = [\dfrac{\partial \mathcal{L}}{\partial b_{N_q -1}}, ..., \dfrac{\partial \mathcal{L}}{\partial b_{0}} ]
\end{equation}
where $\mathcal{L}$ is the inference loss of DNN parametrized by $\vb$.
The naive operation is to directly perform the bit-flip using the gradients obtained in \cref{eqt:bit_grad} and get perturbed bits as: 
\begin{equation}
\label{eqt:naive_BFA}
\hat{\vb} = \vb + \textup{sign}(\nabla_{\vb} \mathcal{L})
\end{equation}
where $\textup{sign}(\nabla_{\vb}\mathcal{L}) \in \{-1,+1\}^{N_q}$. However, since the bit value is constrained between 0 and 1 ($\vb \in \{0,1\}^{N_q}$), flipping the bit as \cref{eqt:naive_BFA} could lead to data overflow. Ideally, the BFA is supposed to follow the truth table in \cref{table:BFA_operation}. Thus, we mathematically redefine the BFA as follows:
\begin{equation}
\label{eqt:mask_generation}
    \vm = \vb \oplus \big(\textup{sign}(\nabla_{\vb}\mathcal{L})/2 + 0.5\big)
\end{equation}
\begin{equation}
    \hat{\vb} = \vb \oplus \vm
\end{equation}
where $\oplus$ is the bit-wise {\tt{xor}} operator. $\vm$ is the mask which indicates whether to perform the bit-flip operation.

\begin{table}[ht]
\centering
\caption{Truth table of Bit-Flip Attack (BFA). $b_i$ is the clean bit and $\hat{b}_i$ is the perturbed bit by BFA. $m$ indicate whether there exist value change between $b_i$ and $\hat{b}_i$. The positive and negative of $\partial\mathcal{L}/\partial b_i$ are represented by 1 and 0 respectively. }
\label{table:BFA_operation}
\begin{tabular}{@{}cccc@{}}
\toprule
$b_i$ & sign($\partial\mathcal{L}/\partial b_i$) & $\Hat{b}_i$ & $m$ \\ \midrule
0 & 1~(+) & 1 & 1 \\
0 & 0~(-) & 0 & 0 \\
1 & 1~(+) & 1 & 0 \\
1 & 0~(-) & 0 & 1 \\ \bottomrule
\end{tabular}%
\end{table}

\subsection{Progressive Bit Search}
\begin{figure}[t]
    \centering
    \includegraphics[width=0.45\textwidth]{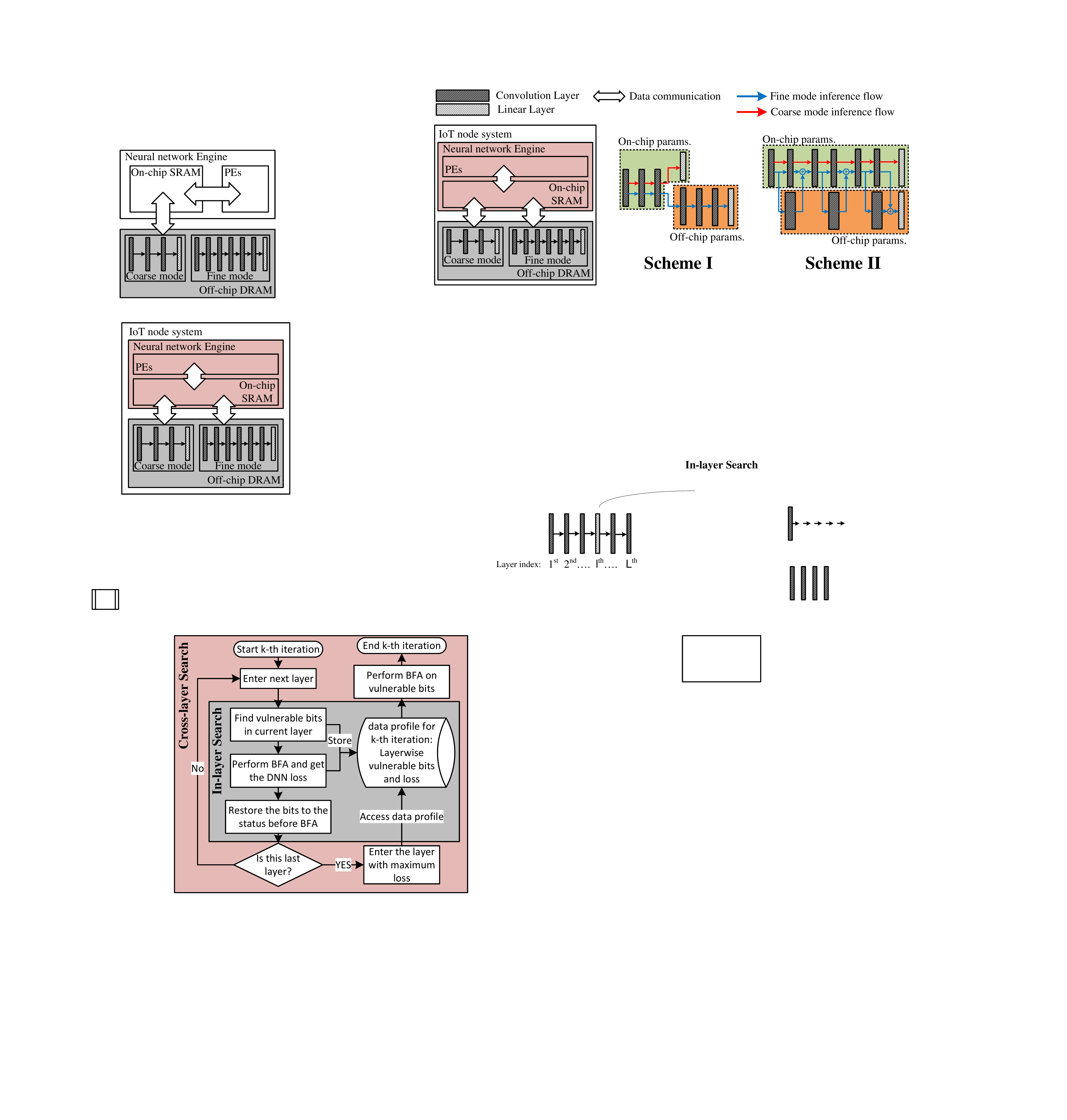}
    \caption{Flowchart to perform Progressive Bit Search (PBS) with in-layer and cross-layer search.}
    \label{fig:BFA_flowchart}
\end{figure}

Rather than performing the BFA upon each bit throughout the entire network, our goal is to perform BFA in a more precise and effective fashion. In this subsection, we propose a method called Progressive Bit Search (PBS) which combines the gradient ranking and progressive search. The proposed PBS method attempts to identify and flip $n_{b}$ most vulnerable bits per BFA iteration ($n_{b}=1$ by default), thus progressively degrading the performance of DNN until it reaches the minimum accuracy or the preset number of iteration. As the flowchart of performing PBS depicted in \cref{fig:BFA_flowchart}, for each attack iteration, the process of bit searching can be generally divided into two successive steps: 1) \textbf{In-layer Search}: the in-layer search is performed through electing the $n_{b}$ most vulnerable bits in the selected layer, then record the inference loss if those elected bits are flipped. 2) \textbf{Cross-layer Search}: with the in-layer search conducted upon each layer of the network independently, the cross-layer search is to evaluate the recorded loss increment caused by BFA with in-layer search, thus identify the top $n_{b}$ vulnerable bits across different layers. The details of each step are described as follows. 

\paragraph{In-layer Search.} 
For the PBS in $k$-th iteration, in-layer searching of the $n_\textup{b}$ most vulnerable bits from $\hat{\textbf{B}}_l^k$ in $l$-th layer is performed through gradient ranking. With the given vectored input $\vx$ and target $\vt$, the inference and back-propagation are performed successively to calculate the gradients of bits w.r.t the inference loss. Then, we descendingly rank the vulnerability of bits by the absolute value of their gradients $\partial \mathcal{L}/\partial b$ and elect the bits whose gradients are top-$n_\textup{b}$, such process can be written as:
\begin{equation}
\hat{\vb}_{l}^{k-1} = \Top_{n_\textup{b}}~\abs{\nabla_{\hat{\textbf{B}}_l^{k-1}} \mathcal{L}\big(f(\vx ; \{\hat{\textbf{B}}_l^{k-1}\}_{l=1}^L), \vt \big) }
\end{equation}
where $\{\Top_{n_\textup{b}}\}$ function returns the pointer pointing at the storage of those elected $n_\textup{b}$ vulnerable bits. Then, we apply the BFA on those elected bits as:
\begin{equation}
\label{eqt:BFA_inlayer_search}
    \hat{\vb}_{l}^{k} = \hat{\vb}_{l}^{k-1} \oplus \vm
\end{equation}
where the mask $\vm$ is generated following \cref{eqt:mask_generation}. Now, with the in-layer search and BFA performed on the $l$-th layer, we have to evaluate the loss increment caused by BFA in \cref{eqt:BFA_inlayer_search}, which can be written as:
\begin{equation}
\label{eqt:inlayer_loss_evaluation}
    \mathcal{L}_l^k = \mathcal{L}\big(f(\vx ; \{\hat{\textbf{B}}_l^k\}_{l=1}^L), \vt \big)
\end{equation}
where the only difference between $\{\hat{\textbf{B}}_l^k\}_{l=1}^L$ and $\{\hat{\textbf{B}}_l^{k-1}\}_{l=1}^L$ are the bits flipped in \cref{eqt:BFA_inlayer_search}.
Note that, those bits flipped to $\hat{\vb}_{l}^{k}$ in \cref{eqt:BFA_inlayer_search} will be restored back to $\hat{\vb}_{l}^{k-1}$ after the loss evaluation is finished.

\paragraph{Cross-layer Search.} As the aforementioned in-layer search can perform the layer-wise vulnerable bits election and BFA evaluation, the cross-layer search evaluates the BFA across the entire network. For the PBS in $k$-th iteration, the cross-layer search first independently conduct the in-layer search on each layer, and generate the loss set as $\{\mathcal{L}_1^{k}, \mathcal{L}_2^{k}, \cdot\cdot\cdot, \mathcal{L}_L^{k}\}$. Then, we could identify the layer-$j$ with maximum loss and re-perform the BFA (without restore) on the bits elected in $j$-th layer, which can be expressed as:
\begin{equation}
\begin{gathered}
\hat{\vb}_{j}^{k} = \hat{\vb}_{j}^{k-1} \oplus \vm \\
s.t.~~ j = \argmax_l~\{\mathcal{L}_l^{k}\}_{l=1}^L
\end{gathered}
\end{equation}
After that, PBS is entered into $k+1$ iteration.
\begin{table*}[ht]
\centering
\caption{BFA on CIFAR-10 with ResNet-20/32/44/56, under various quantization bit-width ($N_\textup{q}$=4/6/8). $N_\textup{flip}$ is the number of bit-flips required (5 trials) to degrade the \textbf{top-1 accuracy below 11\%} with BFA, regardless whether there exists bits flipped back to their original states. For CIFAR-10, top-1 accuracy with random guess is 10\%. $D_{\textup{B}}$ is the hamming distance between clean- and perturbed- binary weight ($D_{\textup{B}} = \sum_{i=1}^{L}\mathcal{D}(\hat{\textbf{B}}_l, \textbf{B}_l)$). The bold number with underline highlight the mismatch between two corresponding $N_\textup{flip}$ and $D_{\textup{B}}$, which indicates there exist even bit-flips on the identical bit/bits. }
\label{table:BFA_CIFAR}
\resizebox{\linewidth}{!}{%
\begin{tabular}{@{}ccccccccccc@{}}
\toprule
 & \multirow{2}{*}{\begin{tabular}[c]{@{}c@{}}Baseline\\ Acc.\end{tabular}} & \multicolumn{3}{c}{$N_\textup{q}=8$} & \multicolumn{3}{c}{$N_\textup{q}=6$} & \multicolumn{3}{c}{$N_\textup{q}=4$} \\  \cmidrule(l){3-11} 
 &  & Acc. & $N_\textup{flip}$ & $D_{\textup{B}}$ & Acc. & $N_\textup{flip}$ & $D_{\textup{B}}$ & Acc. & $N_\textup{flip}$ & $D_{\textup{B}}$ \\ 
 \midrule
Net20 & 92.11 & 92.28 & [7,10,10,12,17] & [7,10,10,12,17] & 91.89 & [8,8,11,12,13] & [8,8,11,12,13] & 91.85 & [7,7,7,8,12] & [7,7,7,8,12] \\
Net32 & 92.77 & 92.32 & [8,9,12,13,31] & [8,9,12,13,31] & 93.09 & [9,10,12,14,23] & [9,10,12,14,23] & 92.31 & [10,12,14,14,17] & [10,12,14,14,17] \\
Net44 & 93.10 & 93.60 & [6,10,11,13,22] & [6,10,11,13,22] & 93.39 & [13,13,15,16,17] & [13,13,15,16,17] & 91.52 & [14,14,15,16,50] & [14,14,15,16,50] \\
Net56 & 92.59 & 93.14 & [16,17,18,22,22] & [16,17,18,22,22] & 93.56 & [16,16,17,20,21] & [16,16,17,20,21] & 92.53 & [9,21,21,\underline{\textbf{23}},24] & [9,21,21,\underline{\textbf{21}},24] \\ \bottomrule
\end{tabular}
}
\end{table*}

\section{Experiments}

\subsection{Experimental setup}

\paragraph{Datasets:}
We take two visual datasets: CIFAR-10 \cite{krizhevsky2010cifar} and ImageNet \cite{krizhevsky2012ImageNet} for object classification task. CIFAR-10 contains 60K RGB images in size of $32\times32$. Following the standard practice, 50K examples are used for training and the remaining 10K for testing. The images are drawn evenly from 10 classes. ImageNet dataset contains 1.2M training images divided into 1000 distinct classes. The data augmentation used in this work is identical to methods in \cite{he2016deep}. Note that, the proposed BFA is performed through randomly draw a sample of input images  from the test/validation set, where the default sample size is 128 and 256 for CIFAR-10 and ImageNet respectively. Then, only the sample input  is used to perform BFA, where the rest data and ground-truth labels are isolated from the attacker. Moreover, each experimental configuration is run with 5 trials to alleviate error caused by the randomness of sampling input. 


\paragraph{Network Architectures and quantization:}
For CIFAR-10, experiments are conducted on series of residual network (ResNet-20/32/44/56)\cite{he2016deep}, where the weights are quantized into 4/6/8 bit-width with retraining. For ImageNet, we choose a variety of famous network structures, including AlexNet, ResNet-18/34/50. Based on our observation, with high bit-width quantizer (e.g., $N_\textup{q}$=8), directly quantizing the pre-trained full-precision DNN without retraining (i.e., fine-tuning) only shows negligible accuracy degradation. Therefore, for fast evaluation of our proposed BFA on ImageNet dataset and its various network structures, we directly perform the weight quantization without retraining before conducting the BFA.

\paragraph{Attack Formulation:}
Traditional attack mostly focuses on attacking DNN by feeding perturbed inputs \cite{goodfellow2014explaining} to the network. Such adversarial attack can be grouped into two major categories: 1) white-box attack \cite{goodfellow2014explaining,madry2018towards}, where the adversary has full access to the network architecture and parameters, and 2) black-box attack \cite{chen2017zoo,papernot2017practical}, where the adversary can only access the input and output of a DNN without its internal configurations. For our proposed BFA, it demands the full access to the DNN's weights and gradients. Thus BFA can be considered as a white box attack. However, we assume that even under white box attack setup, the attacker has no access to the training dataset, training algorithm and hyper parameters used during the training of network. 



\subsection{BFA on CIFAR-10}
Our bit-flip attack is evaluated across different architectures (i.e., ResNet-20/32/44/56) using varying quantized bit-widths (i.e., $N_\textup{q}$=4/6/8) on CIFAR-10 dataset in \cref{table:BFA_CIFAR}. Without BFA, the quantized models show negligible accuracy degradation or even higher accuracy in comparison to their full-precision counterpart. The quantization noise introduced by the weight quantization is considered as a regularization method, which might contribute the accuracy improvement when model training is over-fitting. 


Since CIFAR-10 dataset has 10 different classes of object, degrading the model's accuracy down to 10\% is equivalent to make the model as random output generator. In contrast to adversarial example (e.g., PGD attack \cite{madry2018towards}), our proposed BFA is unable to degrade the network accuracy to 0\%. The reason is adversarial example is an input-specific attack which is designed to misclassify each input separately, while our proposed BFA attempts to misclassify the images from each object category using the identical attacked model. Consequently, the successful BFA would be making the DNN to generate output randomly. Therefore, we report the number of bit-flips $N_{\textup{flip}}$ required to cause the DNN's test accuracy to go below 11\% as the measurable indicator of BFA performance, for CIFAR-10 dataset. 

As the experimental result listed in \cref{table:BFA_CIFAR}, for all the ResNet architecture with varying quantization bit-width, the required number of bit-flips $N_{\textup{flip}}$ to make the DNN malfunction is most likely below 20. Besides $N_{\textup{flip}}$, we take the hamming distance $D_\textup{B}$ between clean- and perturbed-model as another measurable indicator. The intuition behind is our proposed BFA attempts to flip the selected bits without considering its original status. Thus, it exists the probability that some of the bits might be flipped repeatedly with even times. However, the reality is that such back and forth bit-flips rarely happen throughout all the experiments. Under varying quantization configurations, there is no obvious relation between the quantization bit-width and the required number of bit-flips (i.e., robustness of DNN against BFA).


\subsection{BFA on ImageNet}

The summary of evaluation of our attack on ImageNet dataset is presented in table \ref{table:BFA_ImageNet}. We report both baseline and 8-bit quantized network accuracy for four popular image classification architectures on ImageNet. We observe roughly 0.1-0.4 \% reduction in Top-1 classification accuracy after quantizing the network's weights to 8-bits. Since ImageNet dataset has 1000 different classes of objects, a classification accuracy of 0.1\% can be considered as random output. Thus reporting only the number of bit flips $N_{\textup{flip}}$ required to cause the accuracy to degrade to below 0.2\% would be sufficient to prove the attack's effectiveness.

\begin{table}[ht]
\centering
\caption{BFA on ImageNet with various network architecture, under direct 8-bit weight quantization (without retraining). Accuracy (Acc.) is in top1/top5 format. $N_{\textup{flip}}$ is the median number of bit-flips (out of 5 trials) required to \textbf{degrade the top-1 accuracy below 0.2\%}. For ImageNet, top-1 accuracy with random guess is 0.1\%. $D_{\textup{B}}$ is the corresponding hamming distance. Capacity is the number of bits used for weight storage (\# of weights $\times$ 8).}
\label{table:BFA_ImageNet}
\begin{tabular}{@{}ccccc@{}}
\toprule
\begin{tabular}[c]{@{}c@{}}Model\\ (Capacity)\end{tabular} & \begin{tabular}[c]{@{}c@{}}Baseline\\ Acc. $\%$\end{tabular} & \begin{tabular}[c]{@{}c@{}}Quantized\\ Acc. $\%$\end{tabular} & $N_\textup{flip}$ & $D_{\textup{B}}$ \\ \midrule
\begin{tabular}[c]{@{}c@{}}AlexNet \cite{krizhevsky2012ImageNet}\\ (488,806,720)\end{tabular} &
56.55/79.08 & 56.13/78.94 & 17 & 17 \\
\begin{tabular}[c]{@{}c@{}}ResNet-18 \cite{he2016deep}\\ (93,516,096)\end{tabular} & 69.76/89.08 & 69.50/88.98 & 13 & 13 \\
\begin{tabular}[c]{@{}c@{}}ResNet-34 \cite{he2016deep}\\ (174,381,376)\end{tabular} & 73.30/91.42 & 73.13/91.38 & 11 & 11 \\
\begin{tabular}[c]{@{}c@{}}ResNet-50 \cite{he2016deep}\\ (204,456,256)\end{tabular} & 76.15/92.87 & 75.84/92.82 & 11 & 11\\
\bottomrule
\end{tabular}
\end{table}

\begin{figure}[ht]
	\centering
	\captionsetup[subfloat]{farskip=0pt,captionskip=0pt}	
	\subfloat{%
		\begin{minipage}[c][0.42\width]{0.44\textwidth} 
			\centering
			\includegraphics[width=\textwidth]{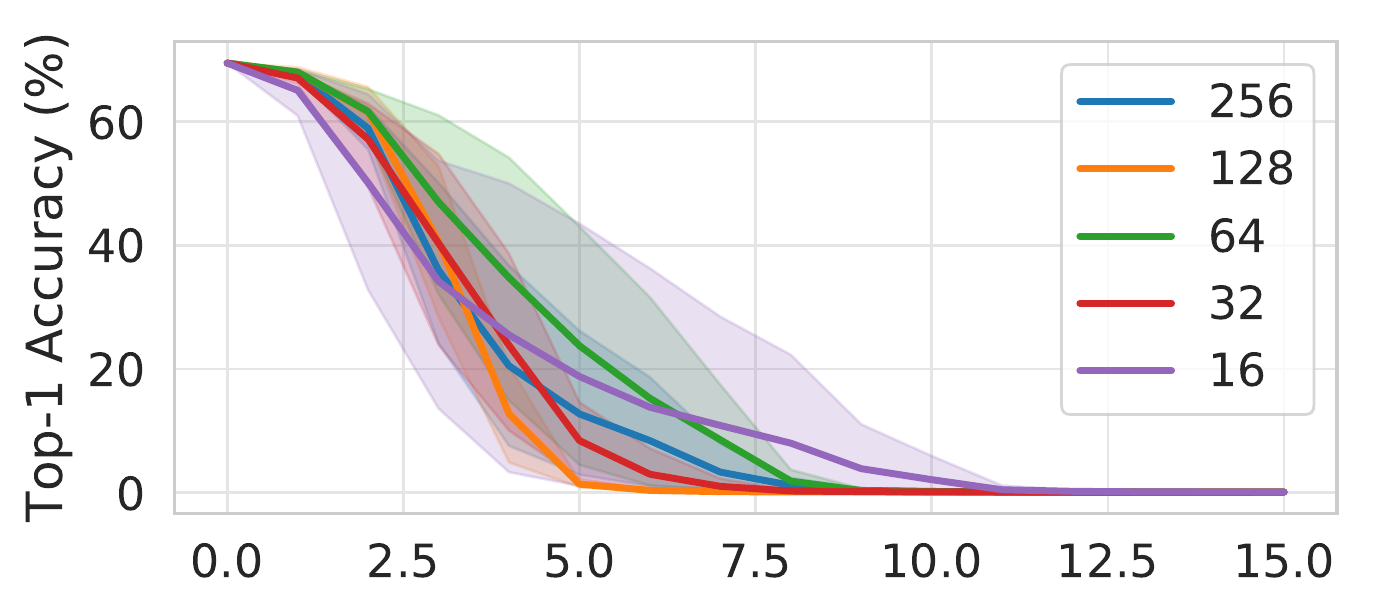}
			\end{minipage}}\\
	\subfloat{%
		\begin{minipage}[c][0.41\width]{0.44\textwidth} 
			\centering
			\includegraphics[width=\textwidth]{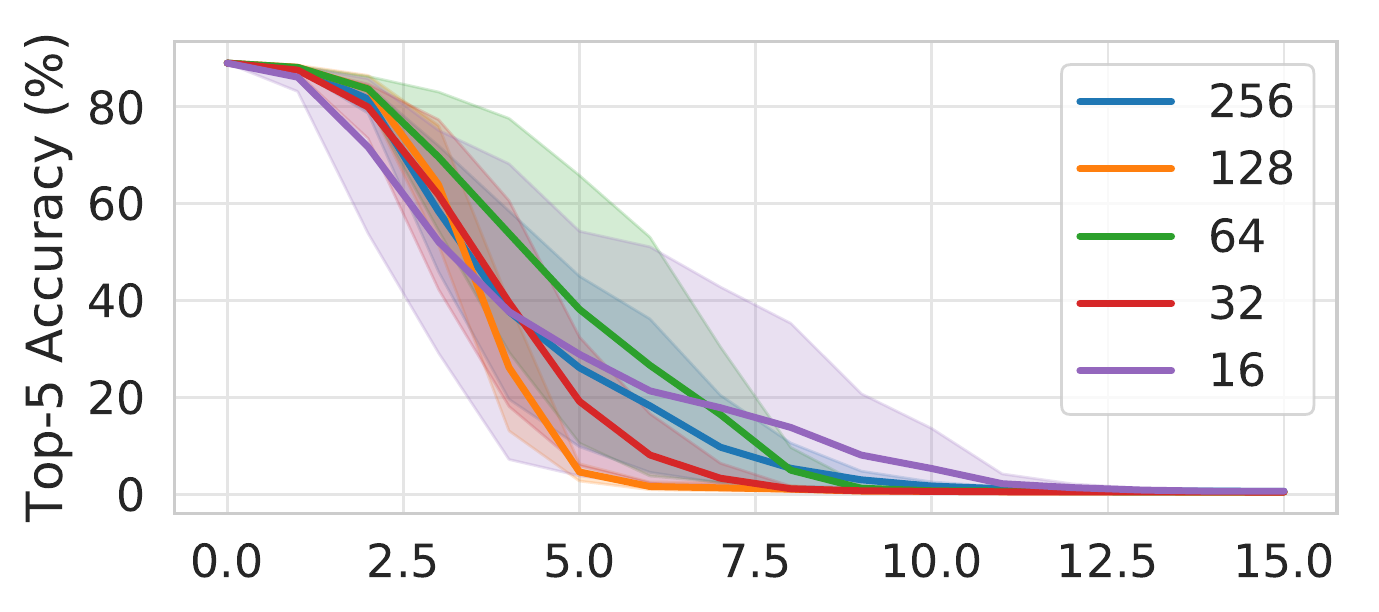} \end{minipage}}\\
	\subfloat{%
		\begin{minipage}[c][0.41\width]{0.44\textwidth} 
			\centering
			\includegraphics[width=\textwidth]{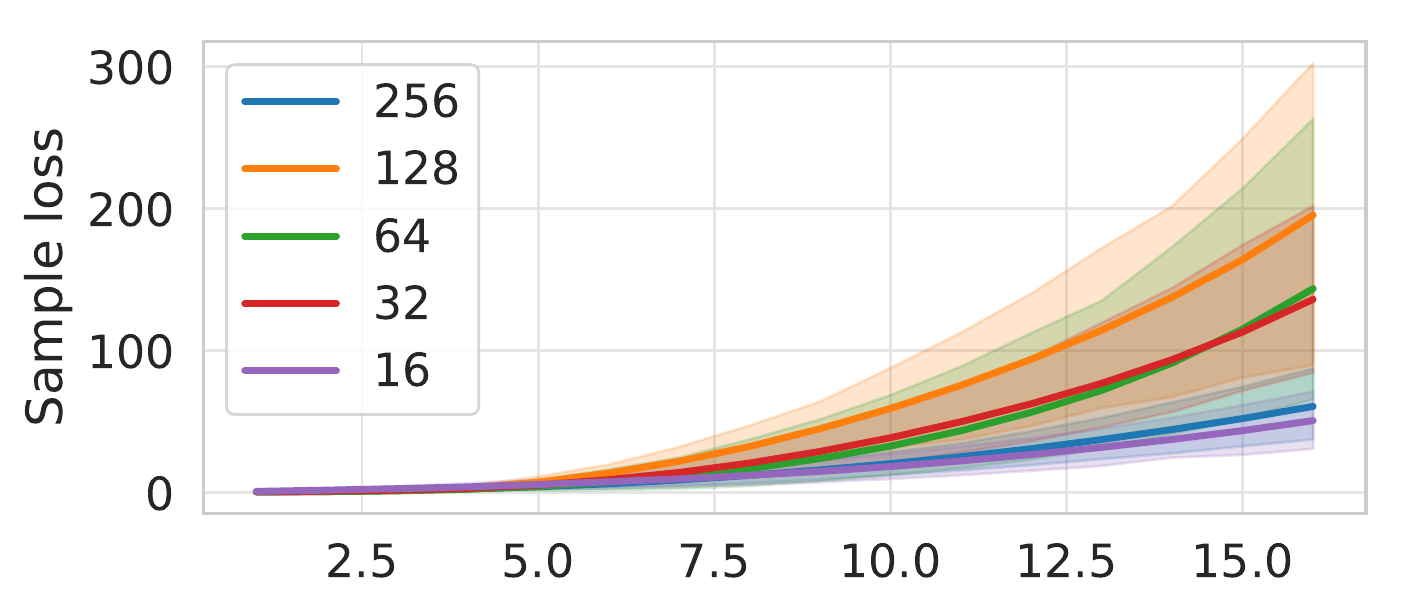} \end{minipage}} \\ 
	\subfloat{%
		\begin{minipage}[c][0.46\width]{0.44\textwidth} 
			\centering
			\includegraphics[width=\textwidth]{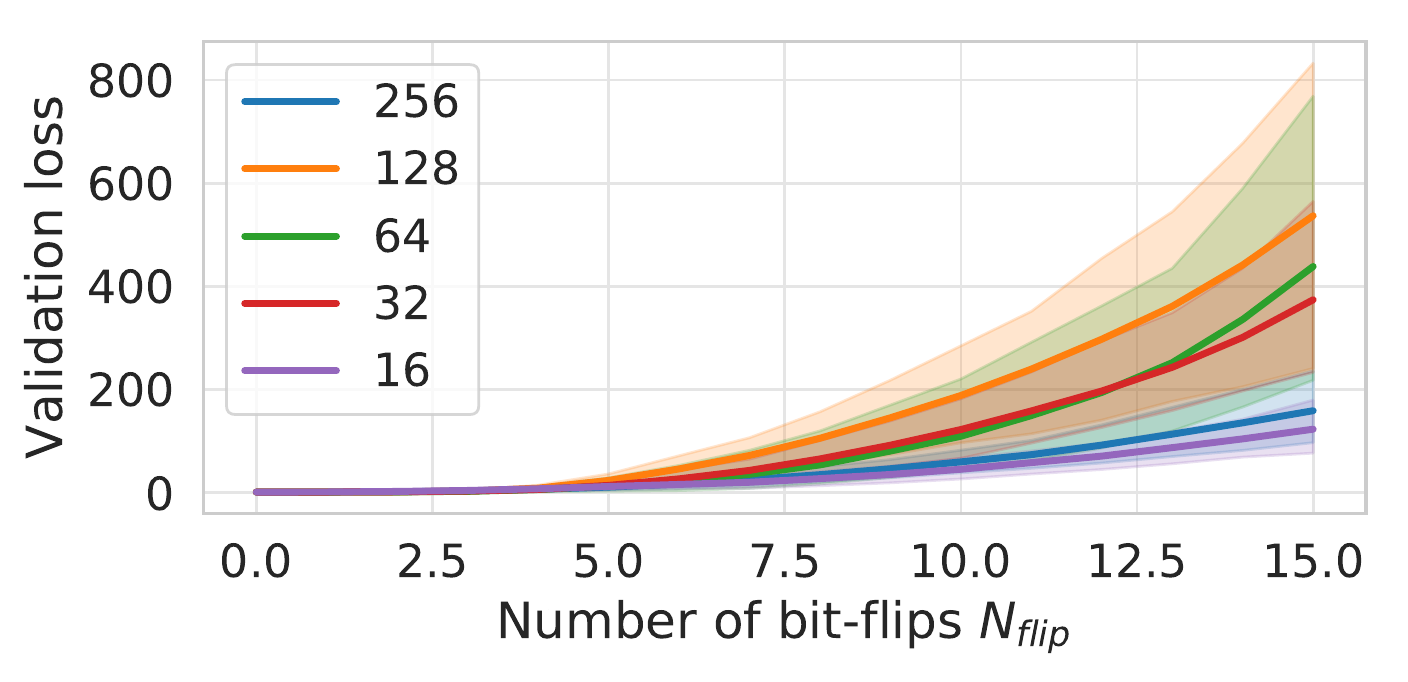} \end{minipage}}
	\caption{The BFA performance of ResNet-18 with various attack sample size (16/32/64/128/256) on ImageNet dataset. Regions in shadow indicates the error band w.r.t 5 trials.}
	\label{fig:imagnet_evolution_batch}
\end{figure}

\begin{figure*}[ht]
	\centering
	\subfloat{%
		\begin{minipage}[c][0.7\width]{0.33\textwidth} 
			\centering
			\includegraphics[width=\textwidth]{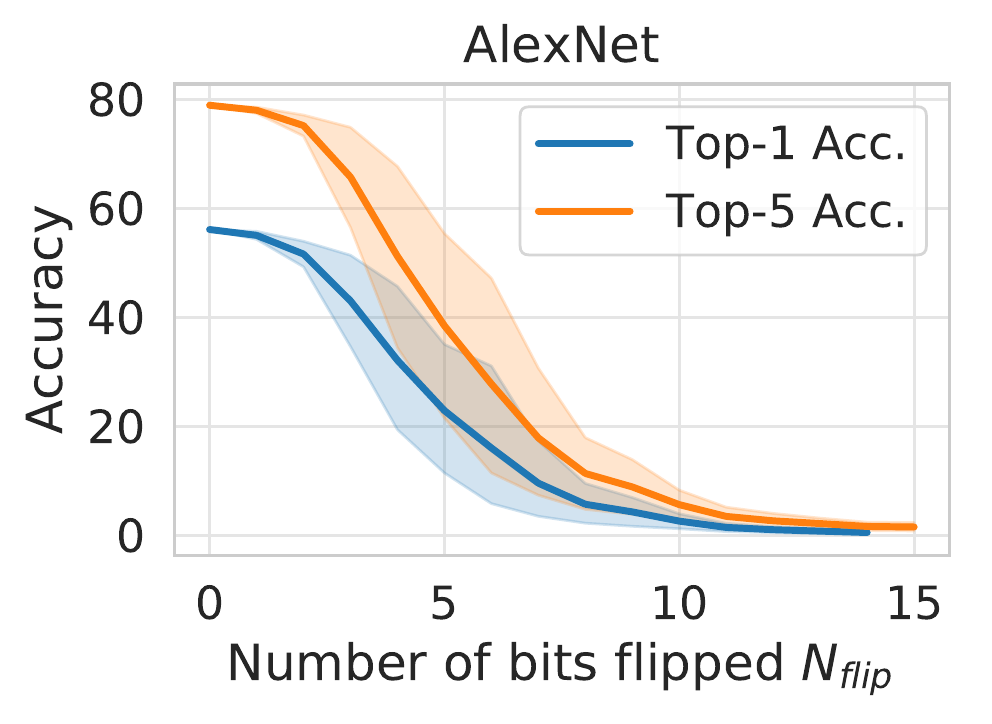}
			\end{minipage}}\hfill
	\subfloat{%
		\begin{minipage}[c][\width]{0.33\textwidth} 
			\centering
			\includegraphics[width=\textwidth]{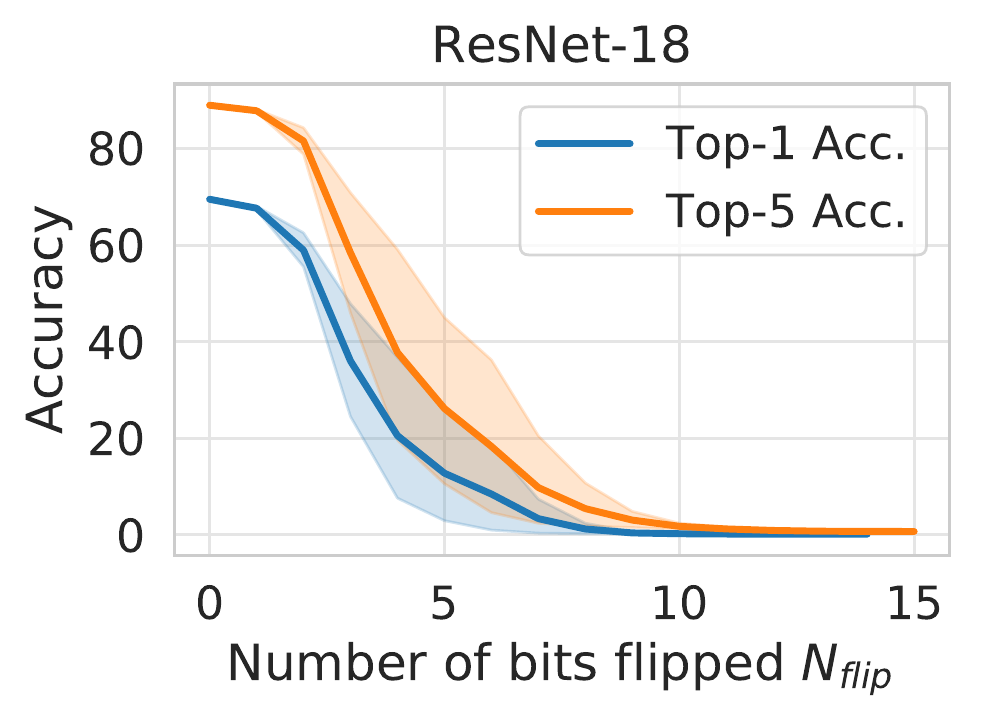} \end{minipage}}\hfill
	\subfloat{%
		\begin{minipage}[c][\width]{0.33\textwidth} 
			\centering
			\includegraphics[width=\textwidth]{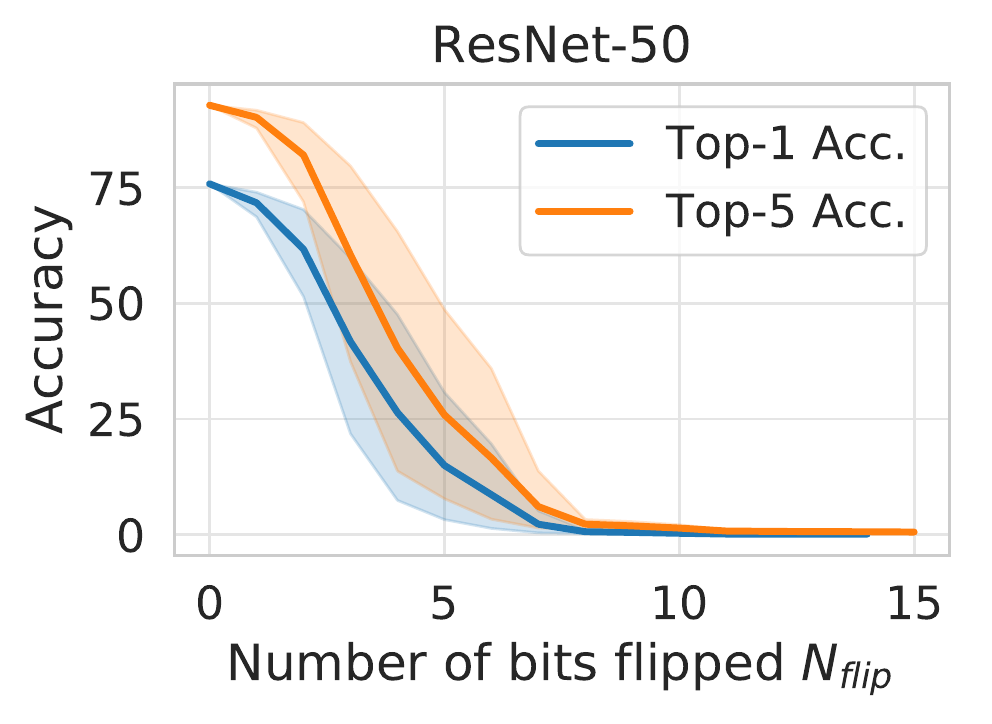} \end{minipage}} \\ 
			
	\subfloat{%
		\begin{minipage}[c][\width]{0.33\textwidth} 
			\centering
			\includegraphics[width=\textwidth]{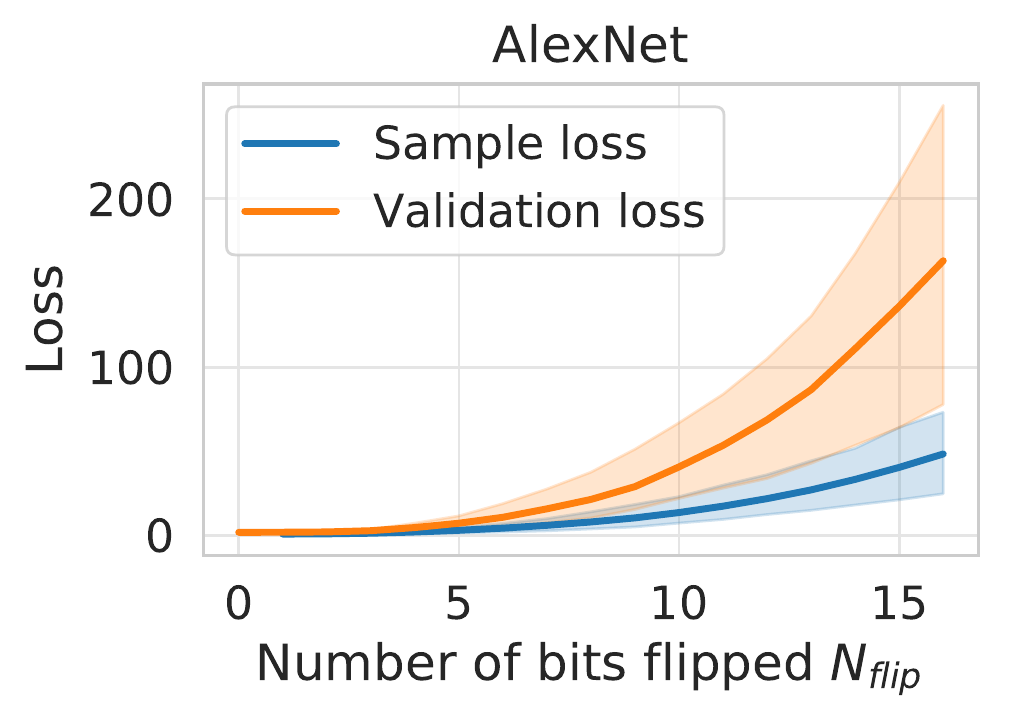} \end{minipage}}\hfill
	\subfloat{%
		\begin{minipage}[c][\width]{0.33\textwidth} 
			\centering
			\includegraphics[width=\textwidth]{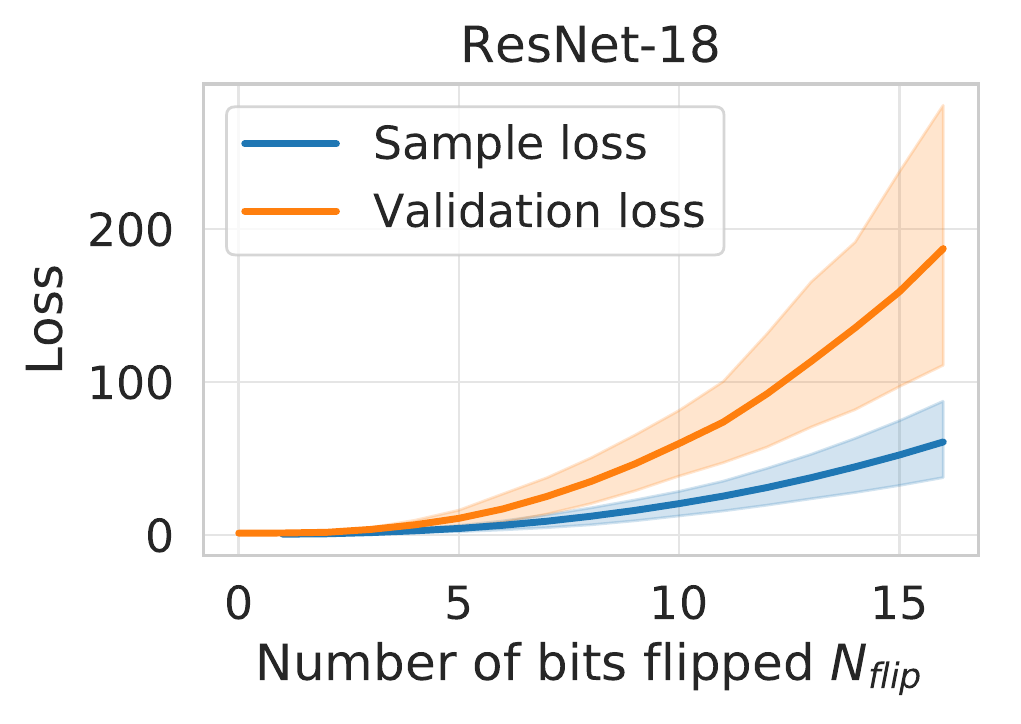} \end{minipage}}\hfill
	\subfloat{%
		\begin{minipage}[c][\width]{0.33\textwidth} 
			\centering
			\includegraphics[width=\textwidth]{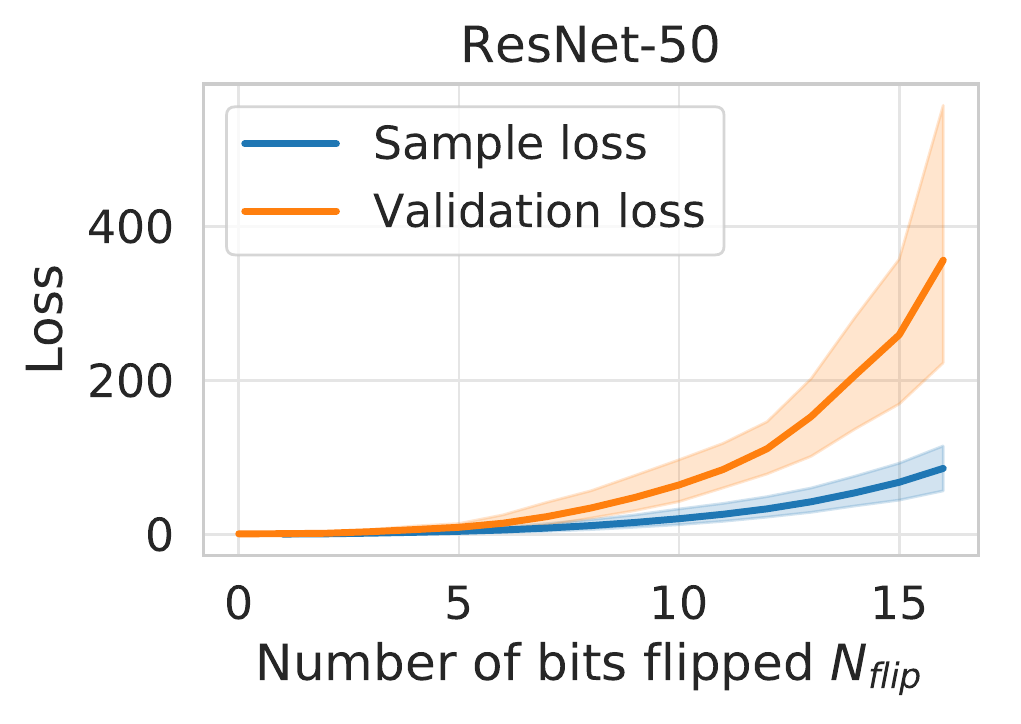} \end{minipage}} 
	\caption{The accuracy (Top1/Top5) and loss evolution curve versus the number of bit-flips ($N_{\textup{flip}}$) under BFA, for AlexNet/ResNet-18/ResNet-50 on ImageNet dataset. The sample size for performing BFA is 256. On each network architecture, we run 5 experiments and the region in shadow indicates the error-band. For all experiments in this figure, there exists no bit flipped multiple times during the attack (i.e., $N_\textup{flip} = D_{\textup{B}}$).}
	\label{fig:imagnet_evolution}
\end{figure*}

For ImageNet, BFA with PBS attack requires only 17 (median of 5 trials ) bit flips out of 480 Million bits to crush AlexNet. However, $N_{\textup{flip}}$ decreases even more as we perform the attack on ResNet architectures. Figure \ref{fig:imagnet_evolution} shows accuracy degradation for ResNet models, which has a much steeper slope than AlexNet. As AlexNet does not have residual connections, which may result in different response to such gradient based attacks. For ResNet networks, as the network parameters keep increasing, it requires lesser number of $N_{\textup{flip}}$ to attack the network. Finally, Our attack makes a ResNet-50 architecture dysfunctional by flipping 11 out of 200 Million bits only. The attack achieves such success by modifying roughly 0.000003\% of the bits to destroy the fully functional DNN. Thus the gravity of DNN parameter's security concern can be summarized as two identical models with 50M similar weights but only a 0.000003\% error in the parameters can generate totally different output values causing a 63\% degradation in test accuracy.

\subsection{Ablation study}

\paragraph{PBS with various sample size.}

In our experiment, we randomly sample a set of input images from the test/validation subset to perform the BFA, which we define it as \textit{attack sample}. Then, we evaluate the effectiveness of the attack on the whole test data set which works as a validation. We opted to perform the validation on the whole test dataset including the random batch that was originally selected for the attack because the sample size is too small compared to the whole test dataset for both ImageNet and CIFAR-10. In this section, we perform an ablation study on the attack sample size. In figure \ref{fig:imagnet_evolution_batch}, We configure the sample size from 16-256 and plotted Top-1 validation accuracy, Top-5 validation accuracy, Sample loss and validation loss respectively.

The performance of the attack based on attack sample size can be ranked as: $S(128)>S(32)>S(256)>S(64)>S(16)$. Even though the effect of sample size does not hinder the attack strength much but with a sample size of 128, our attack requires the fewest bit flips to reach 10\%. On the other hand, with a sample size of 16, the attack strength slightly degrades. Our observation encourages not to select a too large or too small attack sample size. One probable explanation would be if we compute the gradient with respect to large samples, then the attack might fail to properly maximize the loss with respect to every sample. Again, if the sample size is too small then the sample loss may not be representative of the whole test data set.

\begin{figure}[ht]
	\centering
	\subfloat{%
		\begin{minipage}[c][0.7\width]{0.255\textwidth} 
			\centering
			\includegraphics[width=\textwidth]{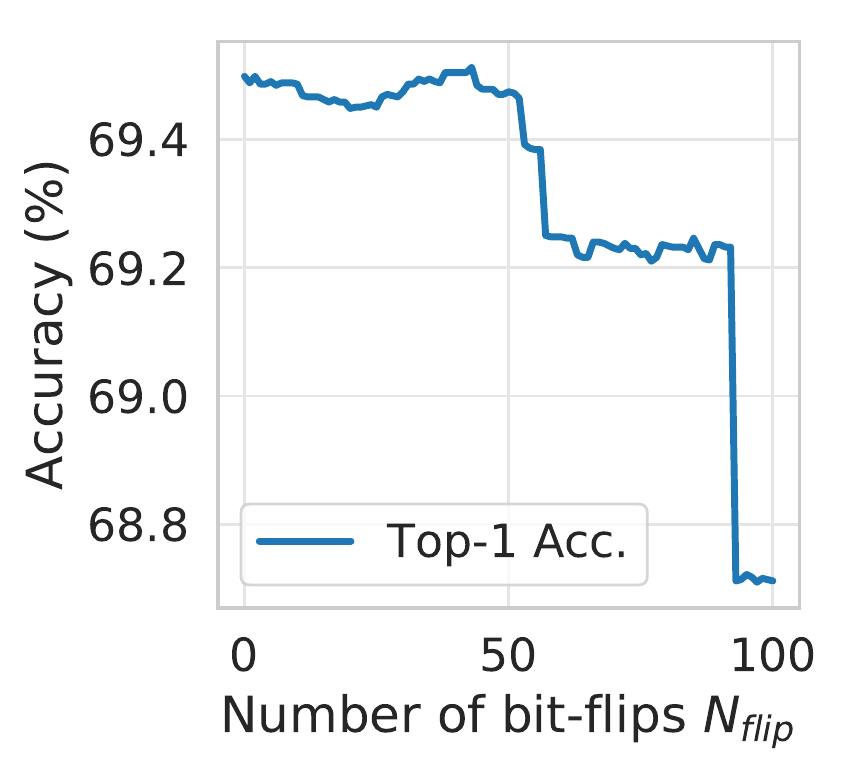}
			\end{minipage}}

	\subfloat{%
		\begin{minipage}[c][0.7\width]{0.24\textwidth} 
			\centering
			\includegraphics[width=\textwidth]{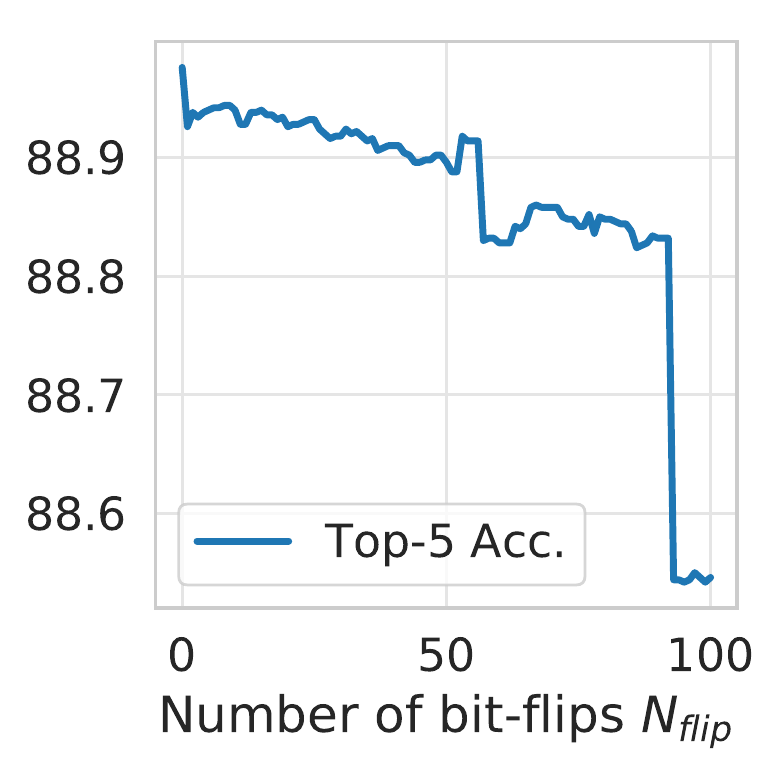} \end{minipage}}
	
	\caption{Randomly flipping bits of a ResNet-18 architecture on ImageNet. Even after flipping 100 random bits the network's both Top-1 and Top-5 accuracy does not degrade significantly.}
	\label{fig:random_attack}
\end{figure}

\paragraph{PBS versus random bit-flips.}
\label{randomflip}

In this section, we perform an ablation study on randomly flipping any bits of a random weight in the network. First, we test random bit flip on a full-precision weight(i.e, floating point) on ResNet-18 model. For floating point weights represented in standard IEEE format, if we change the most significant bits of the exponent section, then the floating-point weight value would change by huge amount. As a result, the trained ResNet-18 Network starts malfunctioning even after just one random bit flip.

Then, we implement the random bit flip on 8-bit Quantized ResNet-18 architecture as shown in figure \ref{fig:random_attack}. It shows that by flipping even 100 random bits, the Top-1 accuracy on ImageNet dataset does not degrade more than 1\%. It demonstrates the need for an efficient bit search algorithm to identify the most vulnerable bits as randomly flipping any bit does not hamper neural network too much. In comparison, our attack algorithm requires just 13 bits out of 93M for ResNet-18 to totally cause the network to malfunction on ImageNet dataset.

\subsection{Comparison to other methods}

Progressive bit search is the very first attack bit searching algorithm developed to malfunction a quantized neural network through perturbation of stored model parameters using row hammer attack. We already showed in previous section that the previous attack algorithms \cite{liu2017fault,breier2018deeplaser} on floating-point model parameters are not efficient. They do not consider that attacking floating point DNN model is as easy as flipping most significant exponent bits of any random weights. Our developed BFA with PBS is the first work that puts emphasis on the need for developing attack algorithms to properly scrutinize the security of DNN model parameters. Our attack can crush a DNN model to demonstrate DNN's vulnerability to intentional malicious bit flips. Further, our algorithm would encourage more future work on both attack and defense front in an attempt to make neural network more resilient and robust.

\section{Discussion}

\paragraph{Why only a few bit flips can cause such destructive phenomena?}

In the analysis of the existence of adversary in deep neural network, Goodfellow et al. \cite{goodfellow2014explaining} concluded that deep neural networks exhibit vulnerability to adversarial examples due to their extreme linearity. The linearity of these models is the reason why they cannot resist adversary. The theory suggests that, with sufficient large input dimension, a network will always be vulnerable to noise injected at any layer. Our proposed BFA with PBS attack also introduces noise at different layers of the DNN. Any noise injected at the intermediate layer will increase as it is multiplied by the input features . 

\begin{table}[ht]
\centering
\caption{Attacking a VGG16 \cite{simonyan2014very} model's only the first and last layer separately on CIFAR-10 dataset. Attacking the first layer is much more effective. The noise injected at the early stages of the network keeps growing as it propagates through the following layers.}
\label{hyp}
\begin{tabular}{ccc}
\toprule
Layer to attack & $N_{\textup{flip}}$ & Accuracy (\%)\\ \midrule
First Conv. layer & 20 & 10.06 \\
Last linear layer & 20 & 84.61 \\
\bottomrule
\end{tabular}
\end{table}

 


For VGG16 network we observed similar phenomena where among the 15 bit flips required to degrade the accuracy to 10 percent, 9 of them are in the first six layers.  Additionally, we confirm this hypothesis of noise propagation across layers by the experiment shown in table \ref{hyp}. We attack the model by freezing all the layers (making them not accessible to the attacker) except the first layer, then we do the opposite by freezing all the layers except the last one. As expected, attacking the first layer achieves higher attack success. However, this linearity theory may be too simple to explain other complex phenomena inside a DNN and may not hold true across different architectures. For example, ResNet architecture which has skip connections, tend to evenly distribute the bit flips across different layers. 

\paragraph{BFA with PBS does not suffer from gradient obfuscation.} 

Generation of adversarial examples in quantized network using straight-through estimator introduces gradient obfuscation \cite{athalye2018obfuscated,lin2018defensive}. Attacking a quantized network becomes tricky as such network shows signs of gradient scattering \cite{athalye2018obfuscated}. In this work, we also used a quantized network which implements a uniform quantizer. However, our network directly uses quantized weights to do the inference after training. We calculate the gradient directly with respect to the quantized weights to avoid gradient obfuscation. Moreover, the performance of BFA against 4,6,8 bits quantized networks proves that the effectiveness of BFA does not degrade due to the presence of a non-differentiable function at the forward path.

\paragraph{Potential Defense Methods.}

In order to defend adversarial examples, most common approach now-a-days is to train the network with a mixture of clean and adversarial examples \cite{goodfellow2014explaining,madry2018towards}. One of the proposed defense methods against BFA would be to train the network to solve Madry's Min-Max optimization problem \cite{madry2018towards}. Their approach called adversarial training minimizes two losses: one from real image and other from adversarial image. Hence, we perform adversarial training using BFA with PBS to minimize two such losses: one computed from the original network and the other computed from the same network with one bit flip for each batch.  

However, unlike adversarial training, such a training method does not help in improving the robustness of the network. Our attack can bypass adversarial training scheme primarily because of a large search space of close to 93M bits. Even if we train the network to be resilient to several bit-flips, there will always remain some bits that will be vulnerable to attack. Another potential defense against BFA can be quantized networks. Again our observation in table \ref{table:BFA_CIFAR}, does not show any co-relation between number of quantization bits with the number of bit-flips required. Thus some of the popular adversarial defense methods \cite{madry2018towards,lin2018defensive} fail against our BFA attack. The above observations make our attack even more threatening for deep learning applications.

\section{Conclusion}

Our proposed attack is the very first work for vulnerable bit search on quantized  neural networks. BFA puts light on why the security analysis for neural network parameters needs more attention. We demonstrate through extensive experiments and analysis that the vulnerability of DNN parameter to malicious bit-flips is extremely severe than anticipated. We would encourage further investigation on both attack and defense front in order to thrive towards developing a more resilient network for deep learning applications.

\bibliographystyle{unsrt}
\bibliography{reference}

\end{document}